\pdfoutput=1

\documentclass[11pt]{article}

\usepackage[final]{acl}

\usepackage{times}
\usepackage{latexsym}

\usepackage[T1]{fontenc}

\usepackage[utf8]{inputenc}

\usepackage{microtype}

\usepackage{inconsolata}

\usepackage{graphicx}

\usepackage{amsmath,amsfonts,amssymb}
\usepackage{xcolor}
\usepackage{colortbl}
\usepackage{microtype}
\usepackage{hyperref}
\usepackage{url}
\usepackage{booktabs}
\usepackage{graphicx}
\usepackage{xspace}
\usepackage{subcaption}
\usepackage{graphicx}
\usepackage{floatflt}

\newcommand{\ssect}[1]{Section~\ref{#1}}
\newcommand{\append}[1]{Appendix~\ref{#1}}

\newcommand{\fig}[1]{Figure~\ref{#1}}
\newcommand{\tbl}[1]{Table~\ref{#1}}

\newcommand{\myparagraph}[1]{\noindent \textbf{#1}}

\newcommand\hc{ \rowcolor{teal!15}}

\newcommand{\quality}{QuALITY\xspace}
\newcommand{\sys}{LLoCO\xspace}

\title{LLoCO: Learning Long Contexts Offline}

\author{Sijun Tan*, Xiuyu Li*, Shishir Patil, Ziyang Wu, Tianjun Zhang,\\ \textbf{Kurt Keutzer, Joeseph E. Gonzalez, Raluca Ada Popa} \\
UC Berkeley \\
\texttt{\{sijuntan,xiuyu\}@berkeley.edu} 
}

\begin{document}
\maketitle
\def\thefootnote{*}\footnotetext{Equal contribution}\def\thefootnote{\arabic{footnote}}

\begin{abstract}
Processing long contexts remains a challenge for large language models (LLMs) due to the quadratic computational and memory overhead of the self-attention mechanism and the substantial KV cache sizes during generation. We propose \sys, a novel approach to address this problem by learning contexts offline through context compression and in-domain parameter-efficient finetuning with LoRA. Our method enables an LLM to create a concise representation of the original context and efficiently retrieve relevant information to answer questions accurately. 
Our approach extends the effective context window of a 4k token LLaMA2-7B model to handle up to 128k tokens. We evaluate our approach on several long-context question-answering datasets, demonstrating that \sys significantly outperforms in-context learning while using $30\times$ fewer tokens during inference. \sys achieves up to $7.62\times$ speed-up during inference and $11.52\times$ higher throughput during finetuning, substantially reduces the cost of long document question answering. This makes it a promising solution for efficient long context processing.\footnote{Our code is publicly available on \url{https://github.com/jeffreysijuntan/lloco}.}
\end{abstract}

\section{Introduction}
\label{sec: intro}

Large language models (LLMs) have demonstrated remarkable performance across a wide range of tasks~\citep{touvron2023llama, jiang2023mistral}. Many of these tasks require LLMs to understand and reason about long contexts. For instance, document question answering is one of the most common applications of LLMs, where the model is presented with a document as context and asked to respond to related questions or summarize the text. These documents may range from lengthy articles to entire books, potentially surpassing the context window limit of LLMs. Consequently, there is a growing trend in both academia and industry to enhance LLMs' capability to process longer contexts effectively~\citep{chen2023extending, jiang2023mistral, peng2024yarn, chen2024longlora}. This need has driven innovation among LLM providers like OpenAI and Anthropic to develop models that can handle increasingly lengthy texts consisting of several thousands of tokens.

Despite the impressive progress made by the LLM model providers, scaling these models to adeptly manage extended contexts remains a formidable challenge, both technically and financially. Due to the self-attention mechanism, transformer-based LLMs incur a quadratic computational and memory overhead as sequence length increases. Many long-context tasks require reusing the same context many times, which incurs extra latency overhead and substantial costs, as most commercial LLMs operate on a pricing model that is directly tied to the number of tokens processed. For example, a single inference run with a document that has 100k tokens would take 1.5 USD on Claude 3 Opus\footnote{\url{https://www.anthropic.com/api}} and 1 USD on GPT-4-turbo\footnote{\url{https://openai.com/pricing}}. 

\begin{figure*}[h]
    \centering
    \includegraphics[width=1.0\textwidth]{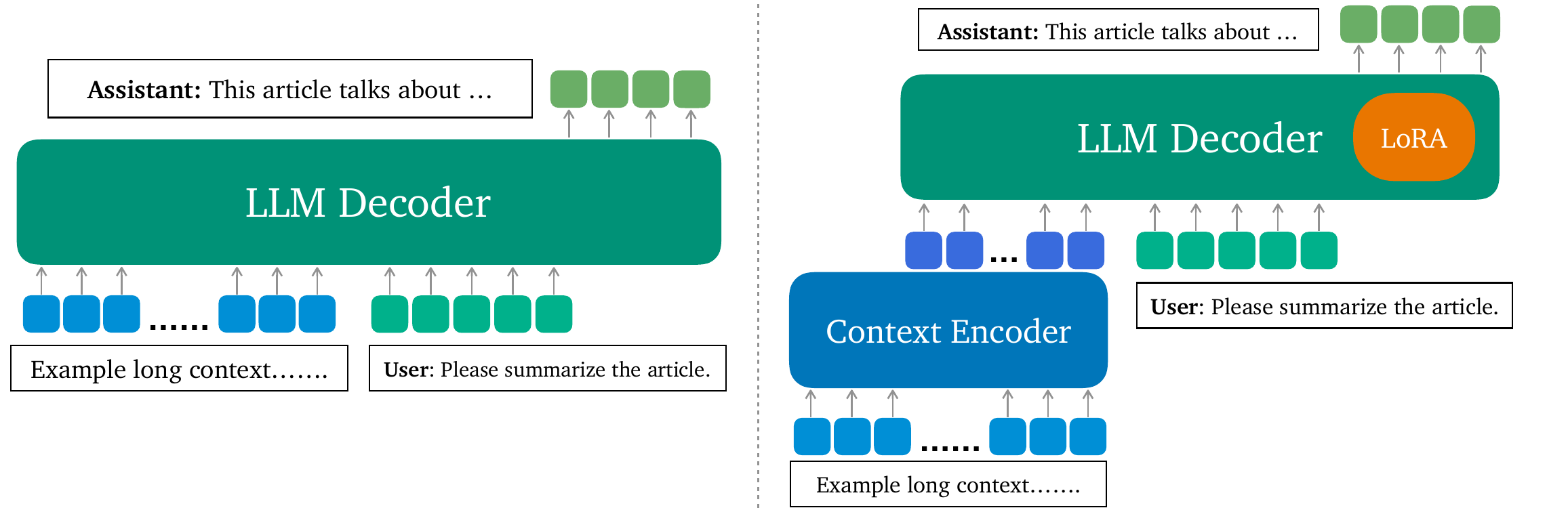}
    \caption{The architecture of regular LLM (left) vs \sys (right). In regular LLMs, long contexts are appended directly to the prompt. In contrast, \sys first processes these contexts through a context encoder. The resulting summary token embeddings are then prepended to the LLM’s prompt, which are significantly shorter. \sys instruction finetunes on embeddings of targeted document groups using a LoRA module. This aligns the LLM’s embedding space with the summary embeddings while keeping both the LLM and context encoder unchanged.}
    \label{fig:lloco-arch}
    \vspace{-14pt}
\end{figure*}

Orthogonal to the exciting efforts of expanding the context window limit, our study introduces an innovative strategy to tackle the long context challenge. We propose a method where context information is condensed offline through finetuning, enabling the model to provide accurate responses during inference with streamlined contextual representations. To illustrate our idea, consider an analogy: \textit{envision an LLM as a student preparing for an exam, where we, the researchers, are the examiners providing study materials and questions}. Traditional in-context learning with full context or Retrieval-Augmented Generation (RAG) resembles an open-book exam, where the LLM has access to all materials while answering questions. In contrast, our approach is akin to a semi-closed-book exam, where the LLM cannot bring the entire book but is allowed to bring a cheat sheet. To excel in the exam, the student must 1) study efficiently to distill a concise yet informative cheat sheet, and 2) effectively retrieve relevant information from the cheat sheet to accurately answer exam questions. Namely, 1) How can we train a model to produce a compact representation of the original context that the LLM can interpret and utilize effectively? 2) How to enable the LLM to proficiently navigate and extract pertinent details from this representation during inference? 

Regarding the first question, we find the formulation closely resembles the existing research direction on context compression. Prior works~\citep{chevalier2023autocomp, mu_learning_2023, ge_-context_2023, yen2024cepe} have proposed compressors that aim to distill the essence of the original texts into compact representations that are aligned with LLMs. Our work primarily focuses on the second question -- we've observed that despite the progress in context compression, LLMs often struggle to accurately read such ``cheat sheets" and tend to hallucinate when applying them to answer queries. To address this issue, we employ in-domain parameter-efficient finetuning directly on the compacted context (cheat sheet) without altering its content, which significantly improves the LLM's ability to accurately extract and utilize information from these compressed representations. In this way, we can steer the LLM to process long context more efficiently and accurately. This approach allows us to extend the effective context window of a 4k LLaMA2-7B model to handle up to 128k tokens. Moreover, we achieve state-of-the-art results that match or even surpass the performance of a LLaMA2-7B-32k model with full context on long context benchmarks, while using $30 \times$ fewer tokens.

This insight has led us to introduce \sys, a pipeline that learns contexts offline through the combination of context compression and parameter-efficient finetuning. It consists of three stages: preprocessing, finetuning, and serving. First, we preprocess the documents into ``cheat sheets''. Then, we employ Low-Rank Adaptation (LoRA)~\citep{hu_lora_2021} to perform parameter-efficient finetuning on these ``cheat sheets'' in groups. 
For serving system design, we use a standard RAG retriever to retrieve the compressed document as well as the most relevant LoRA module, and apply them to the LLM for inference. The contributions of our work could be summarized as follows:
\begin{itemize}
 \item We introduce a novel method for effectively modeling long contexts by combining context compression with instruction finetuning. Our approach extends the context window of a 4k LLaMA2-7B model to handle up to 128k tokens, achieving performance that significantly surpasses in-context learning while using $30 \times$ fewer tokens.

\item We propose \sys, a novel framework that combines context compression, retrieval, and parameter-efficient finetuning. This pipeline could be deployed to significantly speed up and reduce the cost of long document question answering. With the compressed context during inference and instruction finetuning, we demonstrate that LLoCO achives a $7.62\times$ speedup over inference latency, and a $11.52\times$ higher throughput compared to finetuning over the original context.

\end{itemize}

\section{Related work}
\myparagraph{Long-context LLMs} Recently, there have been efforts to increase the LLMs' context window sizes efficiently with continual pretraining or finetuning. One line of work focuses on scaling the Rotary Position Embeddings (RoPE)~\citep{su2021roformer}, achieving longer contexts up to $128$k~\citep{chen2023extending, chen2024longlora, peng2024yarn}. Mistral~\citep{jiang2023mistral} proposes sliding window attention that only attends to part of tokens from the previous layer, reducing compute and enabling pretraining with long-context to $30$k. Nevertheless, as the auto-regressive generation in LLMs is largely memory-bound~\citep{kwon2023efficient}, storing the KV cache of longer contexts slows down the inference and requires large GPU VRAMs. 

\myparagraph{Context compression} A closely related topic is context compression, which aims to train a general compressor that can compress any input prompts. GIST~\citep{mu_learning_2023}, AutoCompressor~\citep{chevalier2023autocomp}, and ICAE~\citep{ge_-context_2023} finetune LLMs in a ``soft prompt-tuning" manner, either applying specific regularization in attention masks or utilizing dedicated ``memory tokens" to compress contexts into embeddings with significant shorter lengths. LLMLingua family~\citep{jiang-etal-2023-llmlingua, jiang-etal-2023-longllmlingua, wu2024llmlingua2} proposes a question-aware 
framework to compress prompts in natural languages for black-box LLM APIs. Another line of work employs KV cache compression by eviction~\citep{zhang2023h_2o, xiao2024streaming, li2024snapkv, tang2024quest}, which only keeps informative keys and values for generation during inference,
or quantization~\citep{sheng2023flexgen, liu2024kivi, hooper2024kvquant}. 
However, those previous approaches aim to compress any inputs, which usually incurs rapid performance drops as the compression ratio exceeds a certain limit (e.g. $4\times$), especially for out-of-distribution texts. Additionally, these methods may require custom kernel implementations to achieve real-time acceleration, limiting their practical use. In contrast, our work focuses on extreme compression of in-distribution documents, achieving up to 30× compression without the need for custom kernels.  

\myparagraph{Retrieval-augmented Generation}
Retrieval enhances the performance of LLMs
in knowledge-intensive tasks such as question answering with long documents or in open-domain. RAG techniques are effective both in finetuned scenarios and when applied to off-the-shelf LLMs~\citep{guu2020realm, lewis2020retrievalaugmented, zhang2024raft}, and have seen many advancements recently, emphasizing improved external knowledge integration, broader query support, and enhanced content clarity~\citep{jiang2023active, schick2023toolformer, asai2023selfrag}. Despite the improved performance from retrieval, challenges still remain in terms of runtime efficiency~\citep{mallen2022trust} and effective filtering of irrelevant parts~\citep{shi2023large, xu2024recomp}. Our proposed method opens up new opportunities for capturing important information while ensuring efficient generation.

\myparagraph{Parameter-efficient Finetuning}
A line of methods~\citep{hu_lora_2021, lester_power_2021, shi2023toast, liu2024dora} freezes the majority of the model weights and only updates a small subset to finetune large models efficiently.
LoRA~\citep{hu_lora_2021} is one of the most widely adopted techniques, and recent advancements~\citep{sheng_s-lora_2023,chen_punica_2023} have focused on enhancing the system performance for deploying LoRA adaptors. In particular, \citet{sheng_s-lora_2023} has achieved the deployment of thousands of LoRA instances on a single NVIDIA A100 GPU. \sys, which utilizes LoRA for finetuning, could benefit from those improvements to deploy LoRA adaptors efficiently in long context scenarios. 

\section{Method}
\label{sec:method}

\subsection{Architecture Overview}

Figure \ref{fig:lloco-arch} illustrates our proposed \sys architecture, which consists of two components: a context encoder and an LLM decoder. In a typical instruction-finetuned LLM, the prompts can be categorized into system, user, and assistant prompts. The system prompt contains task instructions and rules the LLM should follow, as well as relevant contexts. The user prompt is a question asked by the user, and the assistant prompt is the answer generated by the LLM. The context information in the system prompt can include the user's interaction history or documents related to the user's question. These contexts can become very long, potentially surpassing the LLM's context window limit.

To overcome the context window limitation, we propose using a context encoder to compress the original long contexts into a much more compact representation. The context encoder itself is a language model that takes a sequence of tokens as input and outputs a sequence of token embeddings. These output token embeddings, which we call summary embeddings, should be significantly shorter than the original context. The summary embeddings are then prepended to the LLM decoder and serve as the LLM's system prompt. The user's prompt is processed normally through the LLM decoder's tokenizer, and the LLM generates answers (the assistant prompt) conditioned on the summary embeddings and user prompts.

In our design, the context encoder can be any model capable of producing a compact representation aligned with the LLM decoder. The summary embeddings can be viewed as pseudo-words in the LLM decoder's text embedding space, representing abstract concepts or summaries. For our experiments, we select AutoCompressor for LLaMA2-7B \citep{chevalier2023autocomp}, a context compressor finetuned on LLaMA2-7B with the dual ability to generate summary tokens from long contexts and perform text completions conditioned on the summary tokens. The compressor groups the document into chunks of 1536 tokens and recursively compresses each chunk to 50 summary embeddings. To ensure alignment between the summary embeddings and the decoder LLM, we choose LLaMA2-7B as our decoder LLM but use the AutoCompressor's finetuned weights as the model weights. 

While there are other available compressors for LLaMA2-7B~\citep{ge_-context_2023, mu_learning_2023}, we find AutoCompressor to be most suited for our intended use cases given that it can 1) support compressing very long context due to its recursive training procedure, and 2) achieve a great compression ratio of 30:1. We consider the construction of context encoders (compressors) to be an important and orthogonal research direction. Developing more performant and universal context encoders in a streamlined fashion can further enhance the efficiency and effectiveness of \sys, denoting crucial future work.

\subsection{Pipeline for Offline Context Learning}

The pipeline of \sys consists of two primary stages: the preprocessing stage and the finetuning stage. We also outline a serving stage for real-world deployment in the system design. 

\myparagraph{Preprocessing} First, we employ a preprocessing stage that leverages our context encoder to process the original documents. Since AutoCompressor is used as our context encoder, we follow its practice of dividing the documents into chunks of 1536 tokens and passing them through the context encoder. The context encoder outputs 50 summary embeddings for each chunk recursively, with each embedding having a dimension of 4096. 

Our preprocessing stage is designed to be extensible, allowing seamless integration into a retrieval-augmented generation (RAG) system for document QA. In a typical RAG system, preprocessing involves building a vector database to index a collection of documents, where documents are chunked into passages, and each passage is associated with a key— a sentence embedding generated by a text embedding model.The design of \sys allows summary token embeddings to be stored alongside the original passages in the vector database, indexed by the same passage key. 

\myparagraph{Finetuning}
During the finetuning stage, we first segment the documents into groups based on their type (e.g., academic papers, news) or the tasks that users want to perform (e.g., QA, summarization). For each group of documents, we perform parameter-efficient finetuning using a LoRA adaptor. The finetuning data can be in-domain instruction pairs provided by the model provider. If such a dataset does not exist, it could also be generated using self-instruct~\citep{selfinstruct} techniques or distilled from a more powerful model like GPT-4. 
At the end of the finetuning process, we obtain a finetuned LoRA adaptor for each group of documents. 
In the vector database system design, each passage entry will include an identifier for the corresponding LoRA module. Additionally, a separate database will be established to store all adaptors.

Concretely, given summary tokens $\mathbf{X}_{m}$ and instruction pairs $\{\mathbf{X}_{q}^{1},\mathbf{X}_{a}^{1}, \dots, \mathbf{X}_{q}^{L}, \mathbf{X}_{a}^{L}\}$ of length $L$ for a document group $g$, we aim to maximize the probability of generating $\mathbf{X}_{a}$ defined as:
\begin{align}
    p(\mathbf{X}_a | \mathbf{X}_{m}, \mathbf{X}_{q}) = \prod_{i=1}^L p_{\theta_g}(\mathbf{X}_{a}^i | \mathbf{X}_{m}, \mathbf{X}_{q}^{i})
\end{align}
where $\theta_{g}$ denotes the LoRA weight for group $g$.

\myparagraph{Serving} We also design a serving stage that can be naturally extended to a RAG system, where we leverage a standard RAG retriever to retrieve relevant documents, but instead prepend the compressed embeddings of these documents to the LLM decoder. Additionally, we apply the most relevant LoRA module to the model. Readers can refer to \append{appendix:serving} for more details.
\section{Experiments}

In the experiment section, we aim to investigate the following aspects of \sys: (1) its effectiveness in comprehending compressed long contexts, (2) the extent to which summary embeddings can preserve essential information, and (3) the efficiency improvements on the associated system costs.

\myparagraph{Datasets}
We select four datasets dedicated to question-answering tasks—\textbf{\quality}~\citep{pang2021quality}, \textbf{Qasper}~\citep{qasper}, \textbf{NarrativeQA}~\citep{narrativeqa}, and \textbf{HotpotQA}~\citep{hotpotqa}—along with one for summarization, \textbf{QMSum}~\citep{qmsum}. All datasets have long documents as contexts. For all the datasets, we use their validation set for evaluation. We follow the official metrics for each dataset. For \quality, we report the exact match (EM) score. For QMSum, we report the geometric mean of ROUGE scores. For the remaining datasets (Qasper, NarrativeQA, and HotpotQA), we report the F1 scores. More details on these datasets can be found in \append{appendix:dataset}.

\myparagraph{Model Configuration} In this study, we consider two base models. The first is the original LLaMA2-7B-chat \citep{touvron2023llama}, which has a context window of 4096 tokens. The second is Longchat-7b-v1.5-32k \citep{longchat}, a finetuned LLaMA2 model with an extended context window of 32,000 tokens via position interpolation. From now on, we will use LLaMA2-7B-4k to refer to the prior model and LLaMA2-7B-32k to denote the latter one. Unless otherwise specified, we set the LoRA rank to 8 for our experiments. All \sys's models are finetuned on AutoCompressor, which is itself finetuned on LLaMA2-7B.

\begin{table*}[t]
\centering
\resizebox{0.9\linewidth}{!}{\begin{tabular}{llclllllll}

\toprule
Setup & Ctx Size & $\tau$ & \multicolumn{1}{l}{QuA} & QAS & QM & NQA & HQA & Avg.\\ 
\midrule
\multicolumn{8}{l}{\textbf{LLaMA2-7B}} \\ 
\midrule
4k w. Original Context & 4k & 1x & 40.45 & 16.67 & 14.62 & 14.42 & 32.47 & 23.44 \\ 
32k w. Original Context & 32k & 1x & 38.88 & 21.72 & 14.58 & 16.76 & 31.58 & 24.70 \\ 
128k w. Original Context & 128k & 1x & 33.89 & 20.38 & 13.88 & 28.22 & 27.69 & 24.81 \\ 
\midrule
\multicolumn{8}{l}{\textbf{Compression Methods}} \\ 
\midrule
LLaMA2-7B-4k w. Retrieval & 4k & 1.6x & 38.73 & 18.29 & 14.33 & 22.28 & 27.95 & 24.31  \\ 
LLaMA2-7B-32k w. Retrieval & 32k & 12.8x & 36.48 & 24.92 & 15.40 & 19.32 & 22.32 & 23.68  \\ 
\begin{tabular}[c]{@{}c@{}}AutoCompressor\end{tabular} & 128k & 30x & 33.51 & 15.03 & 12.53 & 11.66 & 21.01 & 18.75 \\ 
\midrule
\hc LLoCO w. Full Context (ours) & 128k & 30x & \textbf{41.51} & \textbf{29.01} & \textbf{16.68} & \textbf{28.34} & \textbf{37.83} & \textbf{30.67} \\ 
\bottomrule
\end{tabular}}
\caption{Experiment results on long document QA datasets. $\tau$ indicates the compression ratio. For LLaMA2-7B-4k/32k with retrieval, the compression ratio is obtained by dividing the model's context window limit (4k/32k) by the length of the retrieved passages, which is fixed at 2560 tokens. }
\label{table:exp}
\end{table*}

\subsection{Long Document QA} \label{sec:exp1}
To evaluate the effectiveness of \sys on the aforementioned long context datasets, we consider the following scenarios:

\begin{enumerate}
\item \textbf{LLaMA-2-7B-4k/32k/128k with Original Context.} This is a baseline setting where we provide the LLMs with the ground truth document. We truncate the documents if their length exceeds the context window limit. 
\item \textbf{LLaMA-2-7B-4k/32k with Retrieval.} Retrieval is a standard baseline compression method for long document question answering. For each document, we chunk it into passages of 512 tokens and use Contriever~\citep{contriever} to retrieve the top 5 passages from this document and concatenate them to form the context.
\item \textbf{AutoCompressor.} In this setting, we use AutoCompressor~\citep{chevalier2023autocomp} to compress the contexts into summary embeddings and prepend them to the LLM, without performing any in-domain finetuning, which is equivalent to using it straight out of the box. The AutoCompressor compresses a chunk of 1536 tokens into 50 summary tokens, resulting in an effective context window of roughly 128k tokens. We do not truncate the context unless it exceeds this limit.
\item \textbf{\sys (ours).} \sys is our proposed system for long document question answering. For each dataset we evaluate, we perform instruction finetuning using the question-answering pairs from the training set. During both finetuning and inference, we prepend the summary embeddings of the corresponding ground truth document to the LLM. We do not truncate the context unless it exceeds the 128k context window limit.

\end{enumerate}

Our results are summarized in Table~\ref{table:exp}. When AutoCompressor is used without in-domain finetuning, its performance sometimes falls short of the baseline where no context is appended. However, by combining compression and finetuning, \sys significantly outperforms the baseline on all datasets, using $30 \times$ fewer tokens.

For the \quality, Qasper, QMSum, and HotpotQA datasets, most samples fit within the 32k token limit of the LLaMA2-7B-32k model without truncation. Here, \sys does not have a longer context advantage but still uses $30 \times$ fewer tokens than the baseline while achieving better performance. This shows that in-domain finetuning enhances the model's ability to interpret compressed embeddings and answer questions. Detailed sequence length statistics are in Table~\ref{tab:stat} in Appendix~\ref{appendix:dataset}.

For the NarrativeQA dataset, the average document length is 84,770 tokens, exceeding the LLaMA2-7B-32k context limit. \sys compresses these contexts to about 2,600 tokens, enabling comprehensive context utilization for question answering. To assess the impact of context length and ensure fair comparison, we conducted an ablation study with two new configurations: (1) documents truncated to 32k tokens before compression and finetuning \sys on the resulting embeddings, and (2) testing another LLaMA2-7B extended for $128$k context window with state-of-the-art long-context ability from~\citep{fu2024data}. Our results in Table~\ref{table:nqa-comparison} show that \sys performs better than LLaMA-7B at both 32k and 128k context lengths.

\begin{table}[h]
\centering
\resizebox{0.9\linewidth}{!}{\begin{tabular}{lccc}
\toprule
Setup & NQA \\ \midrule
LLaMA2-7B-32k~\citep{longchat} & 16.76 \\
LLoCO-32k & 27.99 \\
LLaMA2-7B-128k~\citep{fu2024data} & 28.22 \\
LLoCO-128k & 28.34 \\
\bottomrule
\end{tabular}}
\caption{Impact of context lengths on NarrativeQA.}
\vspace{-10pt}
\label{table:nqa-comparison}
\end{table}

\subsection{Ablation Study}
\label{sec:abl}
In this section, we present ablation studies comparing \sys under various setups, including finetuning LLaMA with the original context and evaluating different compression ratios. Additional studies, such as exploring alternative context encoders, are provided in Appendix~\ref{appendix:ablation}.

\myparagraph{Finetuning LLaMA with Original Context}
One interesting question is how well LoRA finetuning works over original uncompressed context, and how does that compare to \sys's approach. To investigate this, we conduct additional finetuning experiments on LLaMA2-7B-4k/32k, where we append the original uncompressed context as a system prompt during finetuning. These models are finetuned following the same setup as \sys. We do not explore the impact of finetuning LLaMA2-7B-128k with uncompressed context due to constraints on compute resources.

As shown in~\tbl{table:llama-sft}, both LLaMA-7B-4k and LLaMA-7B-32k exhibit notable performance improvements after finetuning, with increases of $2.88\%$ and $6.62\%$ respectively. Despite these improvements, the performance of \sys remains comparable to that of the finetuned LLaMA-7B-32k model. This finding suggests that finetuning in a compressed context is just as effective as finetuning in the original context. Compared to full-context finetuning, \sys's finetuning step is considerably faster and more cost-efficient due to the use of a significantly shorter context (see \ref{sec:latency}). This makes \sys a more practical finetuning solution for long document question-answering tasks.

\begin{figure}[h]
    \centering
    \includegraphics[width=\columnwidth]{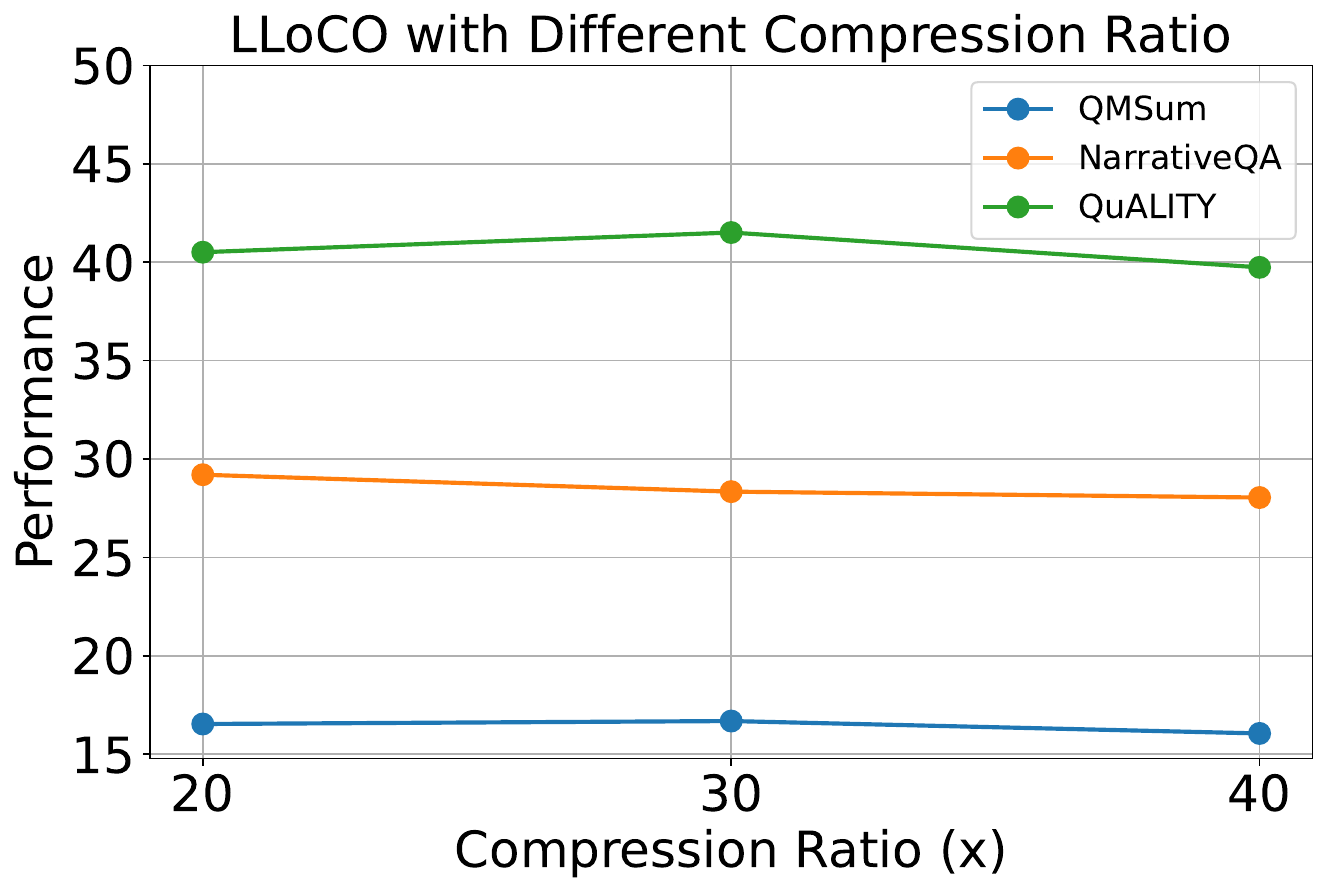}
    \caption{Impact of compression ratio on \sys's performance.}
    \label{fig:comp-ratio}
\end{figure}

\myparagraph{Exploring the Impact of Compression Ratios} In this ablation study, we evaluate \sys under various compression ratios to analyze their effect on performance. The original \sys setup, adapted from AutoCompressor, applies a 30x compression ratio, reducing 1536 tokens to 50 tokens. To explore the impact of different ratios, we also compress 1024 tokens and 2048 tokens into 50 tokens, corresponding to compression ratios of 20x and 40x, respectively. We assess these configurations on the QuALITY, QMSum, and NarrativeQA datasets, with the results presented in Figure~\ref{fig:comp-ratio}.

Our findings suggest that performance remains relatively stable across different compression ratios, with both 20x and 30x ratios consistently outperforming 40x across all datasets. Interestingly, the 30x ratio performs on par with, and occasionally slightly better than, the 20x ratio. We attribute this to the context encoder (AutoCompressor) being originally optimized for the 30x setting, allowing it to maintain high performance even at higher compression rates.

\begin{table*}[h]
\centering
\resizebox{0.9\linewidth}{!}{\begin{tabular}{llllllcll}
\toprule
Setup & \multicolumn{1}{l}{QuA} & QAS & QM & NQA & HQA & Avg. \\ \midrule
LLaMA2-7B-4k & 40.45 & 16.67 & 14.62 & 14.42 & 32.47 & 23.44 \\
LLaMA2-7B-32k & 38.88 & 21.72 & 14.58 & 16.76 & 31.58 & 24.70 \\
\midrule
LLaMA2-7B-4k w. Finetuning & 35.00 & 17.80 & 15.49 & 21.41 & \textbf{41.89} & 26.32 (+2.88\%) \\
LLaMA2-7B-32k w. Finetuning & 40.12 & \textbf{29.71} & 16.36 & \textbf{28.72} & 41.68 & \textbf{31.32} (+6.62\%) \\
\midrule
\hc LLoCO & \textbf{41.51} & 29.01 & \textbf{16.68} & 28.34 & 37.83 & 30.67 \\
\bottomrule
\end{tabular}}
\caption{Comparison of \sys with LLaMA-7B baselines after in-domain finetuning.}
\label{table:llama-sft}
\end{table*}

\subsection{Evaluation on LongBench} 
\label{longbench}
We further evaluate \sys on LongBench~\citep{longbench}, which consists of 6 subtasks. Given that the primary applications of \sys are document question answering and summarization, our evaluation focuses on the SingleDoc QA, MultiDoc QA, and Summarization tasks. There exists overlap with some datasets (e.g. Qasper, QMSum) we evaluated in Section~\ref{sec:exp1}. However, the validation sets differ as LongBench samples a specific subset of examples for each dataset. We have rigorously ensured that LongBench does not feature any questions that overlap with those used in the \sys training, thereby completely removing any risk of data leakage. To achieve the best performance of \sys, we choose LoRA adaptors tailored for each particular category/dataset in our evaluation.

Specifically, for the SingleDoc tasks, we use the NQA LLoCO and Qasper LLoCo for the NQA and QAS tasks, respectively. For MultiField-En (MFE) ~\citep{longbench}, we use GPT-4-turbo to generate question-answering pairs from the training documents and finetune \sys using these pairs. For MultiDoc tasks, we use the combined finetuned \sys from Section~\ref{sec:combined-finetune}, which works well for all tasks. For the Summarization tasks, we create a training dataset by combining the entire QMSum~\citep{qmsum} training data with 5,000 randomly sampled examples from the MultiNews~\citep{multinews} training data. We then finetune \sys on this combined dataset. 

\begin{table*}[h]
\centering
\resizebox{0.9\linewidth}{!}{
\begin{tabular}{lccc|ccc|ccc|c}
\toprule
& \multicolumn{3}{c|}{SingleDoc} & \multicolumn{3}{c|}{MultiDoc} & \multicolumn{3}{c}{Summarization} \\ \midrule
& NQA & QAS & MFE & HQA & WMQA & MSQ & Gov & QMS & MNews & Avg. \\
\midrule
LLaMA-2-7B-4k & 18.7 & 19.2 & 36.8 & 25.4 & 32.8 & 9.4 & 27.3 & 20.8 & 25.8 & 24.0\\
LLaMA-2-7B-32k & 16.9 & \textbf{27.7} & \textbf{41.4} & 32.5 & 20.6 & 9.7 & \textbf{30.8} & 22.7 & 26.4 & 25.4\\
\hc LLoCO & \textbf{23.1} & 26.1 & 26.3 & \textbf{46.2} & \textbf{35.6} & \textbf{27.3} & 17.6 & \textbf{23.4} & 25.0 & \textbf{27.8}\\ \bottomrule
\end{tabular}}
\caption{Evaluation results on LongBench for SingleDoc, MultiDoc and Summarization. Numbers for LLaMA-2-7B-4k/32k are taken from the official LongBench's repository~\citep{longbench-github}}
\label{tab:your_label_here}
\end{table*}

Our evaluation results show that \sys outperforms the baselines on 5 out of the 9 datasets. In particular, \sys excels in the MultiDoc QA task, achieving much better results on all three datasets. \sys also demonstrates comparable performance in two datasets (MultiNews, Qasper), but falls short in the GovReport~\cite{govreport} and MultiField-En datasets. The GovReport dataset is a summarization task that requires the model to generate a one-page summary. We found that \sys does not perform well when the generated content is lengthy, which currently is a limitation of our approach. The lower performance on the MultiField-En dataset could be attributed to the data being out-of-distribution compared to our training data, as this dataset does not provide any training examples. Despite these limitations, the average score of \sys across all datasets is higher than that of the LLaMA2 baseline, highlighting the overall effectiveness of our approach.

\subsection{Needle In A Haystack}
\label{needle}
We further investigate \sys's proficiency to retrieve information across different positions of context window with varying lengths using the renowned Needle In A Haystack task~\citep{needle}. Tailoring this to our pipeline, we randomly select a long article exceeding 32k tokens from the NarrativeQA dataset as the ``haystack". This article is used to test the \sys model, which has been finetuned on this dataset as discussed in \ssect{sec:exp1}.

\begin{figure*}[h]
    \centering
    \includegraphics[width=.9\textwidth]{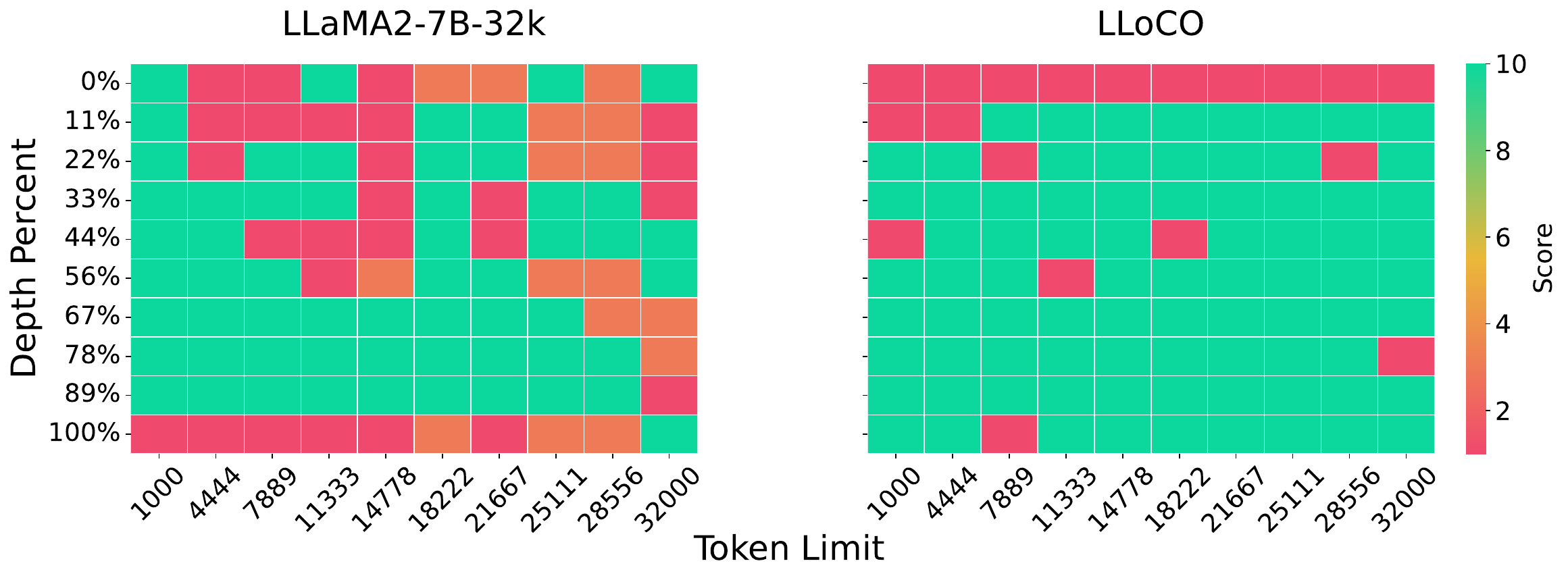}
    \caption{Fixed needle retrieval task. The sampled article ("haystack") starts with "{\tt Mary, ..., a gentle, fashionable girl...}", and a context-relevant needle was curated as "{\tt Mary's favorite fashion designer was Coco Chanel when she was a teenager. Who was Mary's favorite fashion designer when she was a teenager?}"}
    \vspace{-5pt}
    \label{fig:combined}
\end{figure*}

\myparagraph{Single Fixed Needle} Our initial evaluation focuses on a straightforward scenario: the insertion of a consistent fixed ``needle". \fig{fig:combined} shows that our \sys successfully retrieves the needle in $\sim$$80\%$ across all tested context sizes, indicating robust performance with no degradation at longer contexts, while LLaMA2-7B-32k exhibits substantially lower effectiveness in this task.

\begin{figure*}[h]
    \centering
    \includegraphics[width=.9\textwidth]{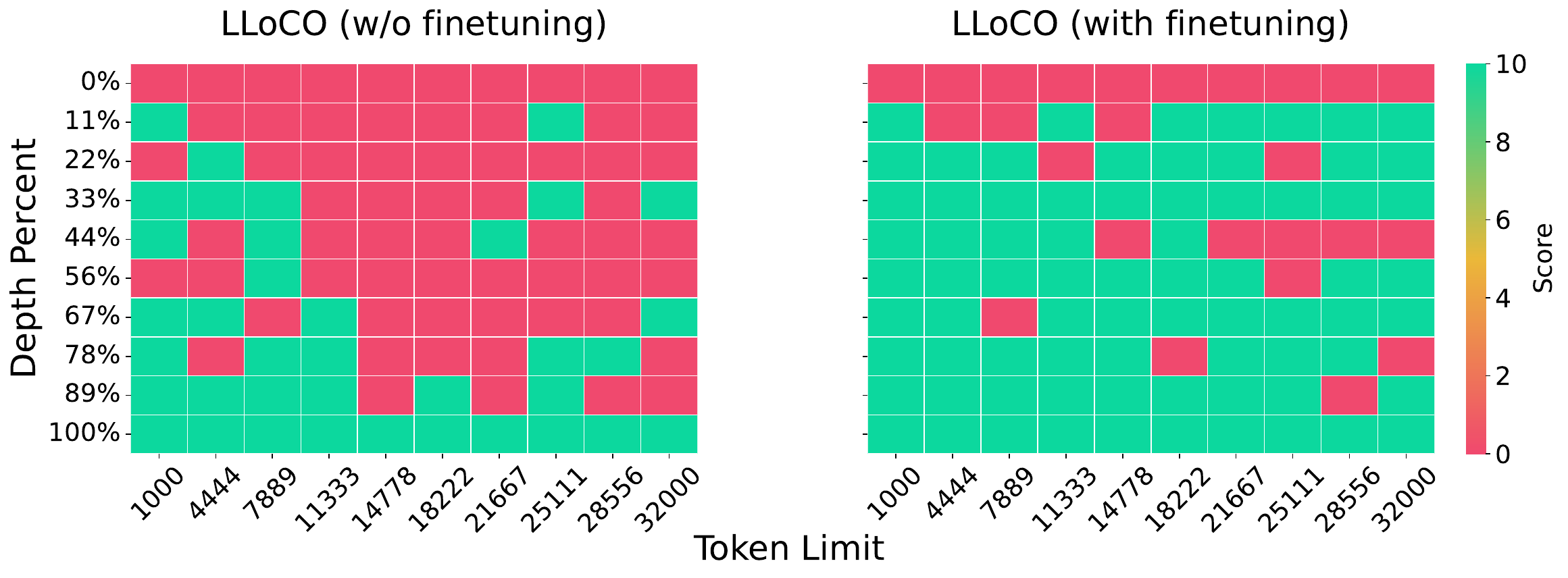}
    \vspace{-10pt}
    \caption{Random needle retrieval with city-word pairs.}
    \label{fig:random_needle}
\end{figure*}

\myparagraph{Random Needles} Additionally, we examine \sys's capability in a more complex scenario by employing a unique ``needle" in each trial. Following~\cite{liu2024world}, we randomly choose a designated city to serve as the needle for each position. We enhance the NQA \sys model through continual finetuning with a synthetic small-scale dataset comprising cities not encountered during evaluations. \fig{fig:random_needle} reveals that although the NQA-finetuned model struggles initially, further finetuning with the \sys pipeline substantially elevates success rates to $\sim$$80\%$. This improvement underscores the efficacy of our method in handling the Needle In A Haystack task.

\subsection{Inference Latency} \label{sec:latency}

\begin{figure}[h]
    \centering
    \includegraphics[width=\columnwidth]{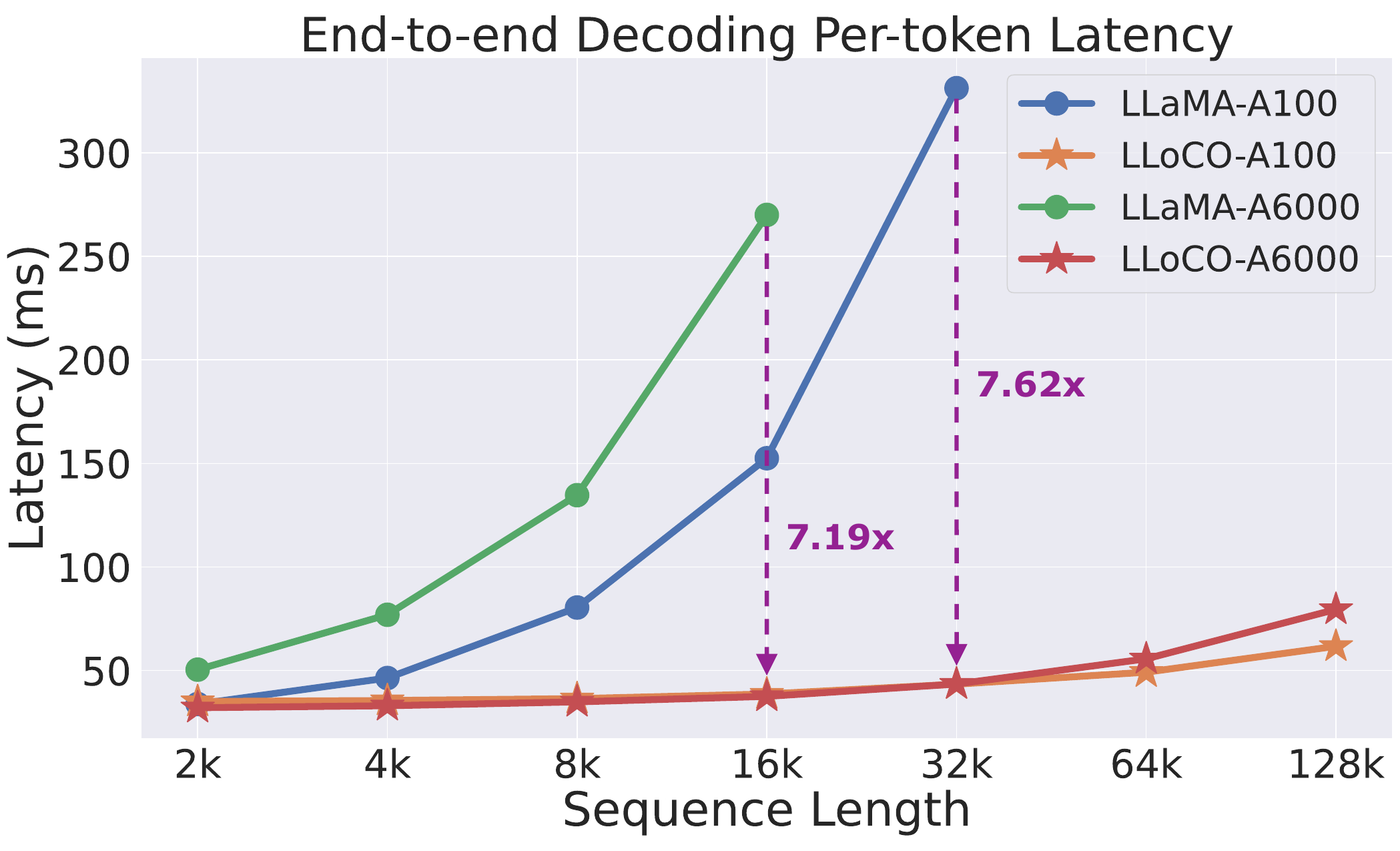}
    \caption{End-to-end decoding per-token latency (ms) on A100 and A6000 GPUs. LLaMA2 without compression runs out of VRAM for sequences of 64k and 128k on A100, and for 32k sequences on A6000.}
    \label{fig:latency}
    \vspace{-10pt}
\end{figure}

In this section, we evaluate the inference latency improvements of our \sys method. The experiments are run on a single A100-80G-SXM GPU and a RTX A6000 GPU, both with a batch size of $1$ and a generation token count set to $16$. We measured the per-token latency with various context sizes. As illustrated in \fig{fig:latency}, \sys realizes speed-ups of up to $7.62\times$ on A100 and $7.19\times$ on A6000 when compared to the LLaMA2-7B baseline without compression, under identical context conditions. While the baseline exceeds GPU VRAM limits for sequences longer than $32$k tokens, \sys maintains efficient generation for sequences up to $128$k tokens. Notably, \sys achieves the baseline's $4$k latency for sequences that are \textbf{16$\times$ longer (64k) on the A100} and \textbf{32$\times$ longer (128k) on the A6000}. Furthermore, \sys can process $32$k sequences on A6000 as fast as baseline's $4$k on A100.

We additionally measure \sys's finetuning throughput improvement over finetuning LLaMA with the original context, where \sys achieves a up to $11.52\times$ higher throughput. The detailed results can be found in Appendix~\ref{appendix:throughput}.

\section{Conclusion}
We propose \sys, a new paradigm that addresses the long-context challenge by preprocessing the context before inference through parameter-efficient finetuning. Our approach extends the effective context window of a LLaMA2-7B model with a 4k tokens context size to handle up to 128k tokens. Evaluations on extensive long-context benchmarks show that \sys significantly outperforms in-context learning while using $30 \times$ fewer tokens during inference, making it a promising solution for efficient long-context processing. 


\section{Limitations}
One limitation of our work is that the context encoder is tied to a specific LLM model. It compresses the original context into a representation aligned with the model. Therefore, for a different model, a separate context encoder must be trained to ensure alignment. Training a context encoder such as AutoCompressor for LLaMA2-7B is costly, requiring about 15 billion tokens of data. On the other hand, studies such as \citet{Morris2023TextER} suggest that original texts can be recovered from sentence embeddings, indicating the possibility of constructing context embeddings that are easily and universally recoverable. Developing a model-agnostic context encoder that can adapt to various models without additional training poses an important challenge and a promising area for future research.

In \sys's pipeline, we separate documents into groups and finetune a LoRA adapter for each group. In a standard RAG pipeline, top $k$ documents relevant to the user's query are retrieved. However, these $k$ documents may come from different groups, each corresponding to a different LoRA adapter, and we can apply only one adapter at a time, which is a limitation. An intriguing future direction would be to investigate the composition of multiple LoRA adapters simultaneously to further enhance QA performance.

There are differences in the effectiveness of \sys across various datasets and tasks. As shown in Section~\ref{longbench} and \ref{needle}, our performance on GovReport and MultiField-En does not surpass the baselines, and we do not achieve perfect results on the Needle-In-A-Haystack task. We attribute these issues to the limitations of the training data quality and the capabilities of the context encoder and LLaMA2 base model. We believe that improving these components will address these shortcomings.

\myparagraph{Acknowledgement}
We thank Michael Luo, Sheng Shen, Chenglei Si and Siyuan Zhuang for their helpful comments and discussion. We also thank BAIR, Berkeley Deep Drive, Intel Corporation, and NVIDIA for supporting this research, as well as Hyperbolic Labs\footnote{\url{https://www.hyperbolic.xyz/}} for providing the AI infrastructure for our experiments.

\bibliography{references}

\newpage
~
\newpage
\appendix


\section{Extended Experimental Settings}
In this part, we provide details of our experimental settings in the main text. For the experiments presented in Sec~\ref{sec:exp1}, we summarize the hyperparameter settings in Table~\ref{tab:hyperparameters_settings}. We use all samples from the train split of each dataset for finetuning, except for HotpotQA, from which we randomly select $20,000$ samples from its train split. All finetuning experiments are conducted with either 4 or 8 NVIDIA RTX A6000 GPUs or A100-80G-SXM GPUs.

\begin{table*}[ht]
\centering
\small
\begin{tabular}{cccccc}
\toprule
\textbf{Hyperparameters} & \textbf{QuA} & \textbf{QAS} & \textbf{QM} & \textbf{NQA} & \textbf{HQA} \\ \midrule
LoRA Rank $r$ & \multicolumn{5}{c}{8} \\ [0.2cm]
Optimizer & \multicolumn{5}{c}{AdamW} \\ [0.2cm]
Weight Decay & \multicolumn{5}{c}{0.00} \\ [0.2cm]
Learning Rate & \multicolumn{5}{c}{2$\times$10$^{-5}$} \\ [0.2cm]
LR Scheduler & \multicolumn{5}{c}{cosine annealing} \\ [0.2cm]
Batch Size &  8 & 8 & 8 & 32 & 32 \\ [0.2cm]
Warmup Ratio & \multicolumn{5}{c}{0.04} \\ [0.2cm]
Epoch & 3 & 3 & 3 & 1 & 3  \\
\bottomrule
\end{tabular}
\vskip 0.1in
\caption{Hyperparameter settings of experiments.}
\label{tab:hyperparameters_settings}
\vskip -0.1in
\end{table*}

\section{More Details on Datasets}
\label{appendix:dataset}
Here we provide more detailed descriptions of the 5 main datasets used for evaluation:
\begin{itemize}
    \item \textbf{\quality} \citep{pang2021quality} is a multiple-choice question-answering dataset over long contexts. It contains 150 articles with an average length of 5000 tokens, with 6737 questions in total. This dataset is particularly suited to evaluate models in a long-context setting.
    \item \textbf{Qasper} \citep{qasper} is a dataset for answering questions about NLP research papers, sourced from the Semantic Scholar Open Research Corpus. It includes various types of questions, ranging from detailed explanations to simple yes/no answers, based on provided documents.
    \item \textbf{QMSum} \citep{qmsum} features summaries and transcripts from meetings across various sectors, including academia and industry, focusing on query-based summarization. Participants are tasked with condensing dialogue transcripts based on specific queries.
    \item \textbf{NarrativeQA} \citep{narrativeqa} is a question-answering dataset derived from complete texts of books from Project Gutenberg and movie scripts from various sources. The challenge here involves generating a concise answer from long and potentially disorderly texts sourced from books.
    \item \textbf{HotpotQA} \citep{hotpotqa}) is a Wikipedia-based dataset that challenges users to derive answers through multi-hop reasoning across several documents, featuring a broad range of questions beyond specific knowledge bases.
\end{itemize}

We additionally provide the statistics of these five datasets in Table~\ref{tab:stat}. 

\begin{table*}[h]
\centering
\begin{tabular}{lccccccc}
\toprule
        & QuA & QAS  & QM & NQA  & HQA \\ \midrule
\# of samples    & 2086 & 1726  & 272 &  5878 & 7405      \\
\# of unique articles   & 115   & 273 & 35 & 115  &  7306 \\ 
Avg word count  & 4122 & 3572 & 10422 & 51925 & 952\\
Max word count  & 5967 & 14640 & 24573 & 353480 & 2694\\ 
Max token length & 9953 & 24205 & 34478 & 556249 & 4000\\
\bottomrule
\end{tabular}
\caption{Statistics of the datasets. The maximum token length is counted using LLaMA2-7B's tokenizer. }
\label{tab:stat}
\end{table*}

\section{Finetuning Throughput}
\label{appendix:throughput}
We assess the finetuning throughput for both \sys and LLaMA2-7B-32k with the original context on the NarrativeQA dataset, which mainly consists of samples exceeding $32$k tokens. We compare the 7B $32$k baseline finetuned on 8 A100s and our \sys finetuned on both 8 A100s and 8 A6000s, all with a per-device batch size of 1 and a global batch size of 32, using 4 gradient accumulation steps. For the 7B $32$k baseline finetuning, samples of more than $32$k tokens were truncated to $32$k. \fig{fig:latency} shows that our \sys achieves $11.52\times$ and $6.82\times$ throughput on A100s and A6000s, respectively. This highlights that \sys not only achieves competitive results, but also improves performance with much greater efficiency compared to finetuning the baseline with full context (as shown in \ssect{sec:abl}).
\begin{figure}[h]
    \centering
    \includegraphics[width=0.9\columnwidth]{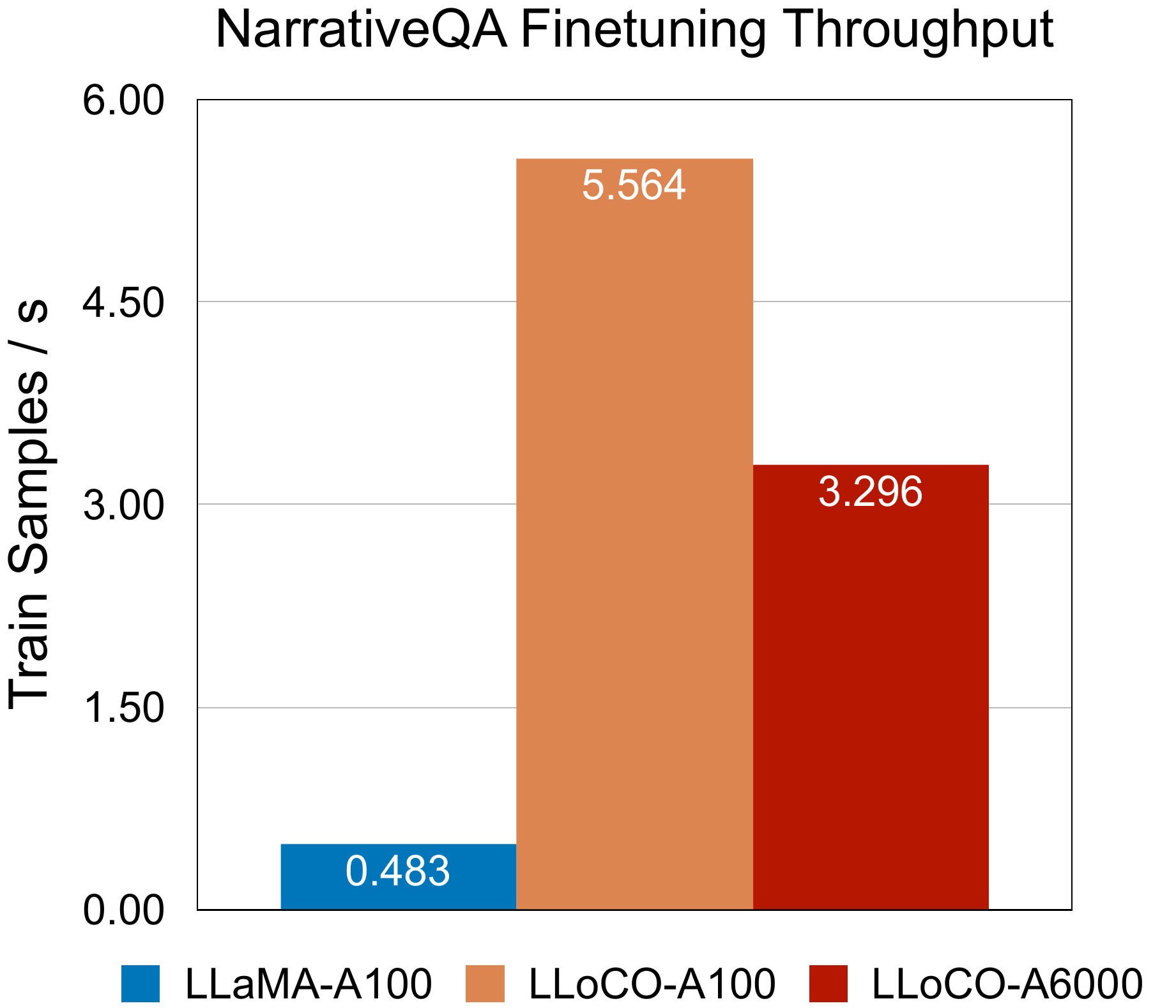}
    \caption{Samples per second when finetuning on NarrativeQA for LLaMA2-7B-32k without compression and \sys.}
    \label{fig:ft-thruput}
\end{figure}

\section{Additional Ablation Study}
\label{appendix:ablation}
\myparagraph{Combined Instruction Finetuning} \label{sec:combined-finetune}
In our previous experiments, we finetuned \sys on each dataset separately. In this ablation study, we investigate \sys's performance when we combine all the training data and finetune a general model. To balance the dataset, we sample 10,000 examples from NarrativeQA and HotpotQA and use all examples from the other three datasets to form our final training set.

\begin{table*}[h]
\centering
\begin{tabular}{llcllllll}
\toprule
Setup & \multicolumn{1}{l}{QuA} & QAS & QM & NQA & HQA & Avg. \\ \midrule
LLoCO w. Separate Finetuning & 41.51 & \textbf{29.01} & \textbf{16.68} & \textbf{28.34} & \textbf{37.82} & 30.67 \\ \midrule
LLoCO w. Combined Finetuning & \textbf{47.07} & 24.95 & 12.77 & 27.93 & 35.55 & 29.65 \\
\bottomrule
\end{tabular}
\caption{Comparison of \sys with in-domain finetuning and combined finetuning.}
\label{table:combined-sft}
\end{table*}

As shown in Table~\ref{table:combined-sft}, \sys with combined finetuning surpasses baseline methods on all datasets except QMSum. The lower performance on QMSum can be attributed to the fact that it is a summarization task that favors longer, more detailed answers, but the combined finetuning process likely shifts the LLM's behavior towards generating shorter, more focused answers, which may not be optimal for the summarization task.

Compared to in-domain finetuning, combined finetuning generally yields lower performance, with the exception of \quality. \quality questions heavily rely on the LLM's reasoning abilities, whereas the other datasets primarily focus on the LLM's capacity to retrieve relevant information from long contexts. We hypothesize that the additional finetuning on diverse datasets enhances the LLM's ability to reason about long contexts, leading to improved performance on \quality.
Overall, the combined finetuning approach demonstrates the potential for knowledge transfer across different tasks. 

\paragraph{Exploring other Context Encoders}
\begin{table}[h]
\centering
\resizebox{0.9\linewidth}{!}{\begin{tabular}{llllllcll}
\toprule
Setup & \multicolumn{1}{l}{QuA} & QAS & QM & NQA \\ \midrule
LLaMA2-7B-4k & 40.45 & 16.67 & 14.62 & 14.42 \\
ICAE & 36.5 & 16.14 & 15.69 & 17.38  \\
\hc LLoCO & \textbf{41.51} & \textbf{29.01} & \textbf{16.68} & \textbf{28.34} \\
\bottomrule
\end{tabular}}
\caption{Comparison of \sys with ICAE~\cite{ge_-context_2023}}
\label{table:icae}
\end{table}

By default, \sys utilizes AutoCompressor as its context encoder. However, \sys offers flexibility in its choice of context encoders and is not limited to a single method. In this ablation study, we evaluate \sys using another state-of-the-art context compression method, ICAE~\cite{ge_-context_2023}, as the context encoder for LLoCO. Since ICAE was trained on contexts up to 4096 tokens, we truncated the inputs to this length, resulting in compressed embeddings of 128 tokens, following the approach outlined in the original paper. 

Our results in Table~\ref{table:icae} show that ICAE performs comparably to LLaMA2-7B-4K across benchmarks, while achieving a 40x compression ratio. This highlights that our proposed pipeline, which combines context compression with PEFT, is not restricted to AutoCompressor and can be generalized to other context encoders.

However, there remains a notable performance gap between ICAE and LLoCO. A key limitation of ICAE is its inability to extend the context window beyond 4K tokens, making it a less suitable option compared to AutoCompressor as the context encoder for our approach.

\section{LLoCO's Serving Stage}
\label{appendix:serving}
\myparagraph{Serving}
In a traditional RAG system, when a user asks a question, the system employs a retriever to fetch the top k relevant documents/passages from the vector database and concatenates them to the prompt to serve as the relevant context.

\begin{figure*}[h]
\centering
\includegraphics[width=0.75\linewidth]{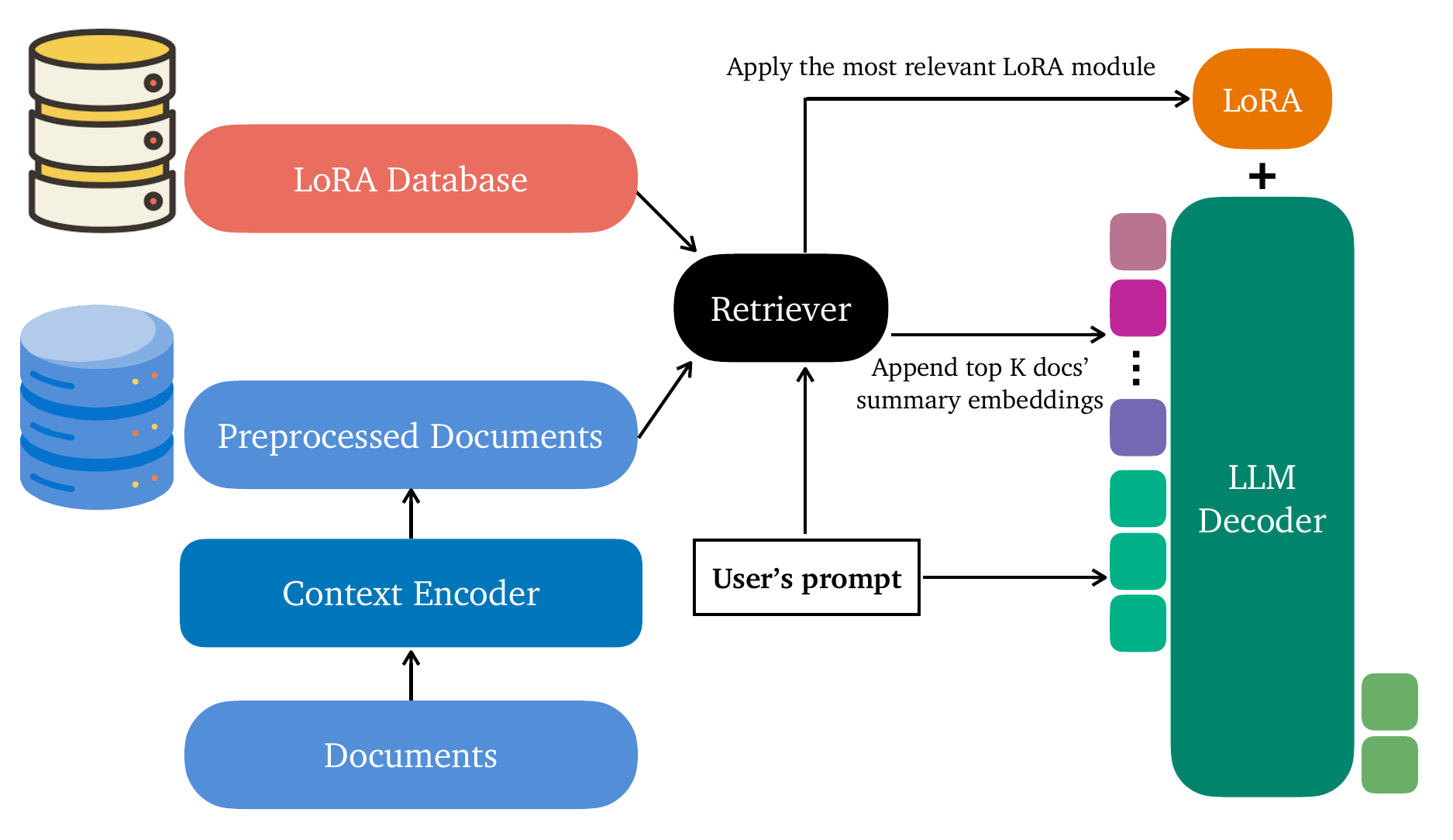}
\caption{The serving stage of \sys's pipeline. }
\label{fig:lloco-pipe}
\end{figure*}

In Figure~\ref{fig:lloco-pipe}, we illustrate \sys's serving pipeline. In a standard RAG, the retriever computes a text embedding for each passage, and use that as the key to retrieve passages relevant to the user query. We use the same text embedding as key, but instead retrieve the compressed token embeddings of the passages, and prepend it to the decoder LLM. Additionally, the system searches for the corresponding LoRA adaptor in the database and applies it to the decoder LLM. Applying a LoRA adaptor incurs minimal overhead to the system cost ~\citep{hu_lora_2021}. By leveraging recent work on serving multiple LoRA adaptors \citep{sheng_s-lora_2023, chen_punica_2023}, we can parallelize LoRA adaptor serving and simultaneously serve requests for documents associated with thousands of LoRA adaptors on a single GPU.

In our current system, we assume that all passages retrieved from the vector database for a given user query belong to the same document group, allowing us to use a single dedicated LoRA adaptor. However, \sys can be extended to handle cases where the retrieved passages span multiple document groups. For instance, it is feasible to design algorithms to dynamically select the most relevant LoRA adaptor based on the majority of the retrieved documents or weight the contributions of different adaptors by their relevance to the query~\citep{muqeeth2024learning}. Another interesting and orthogonal research direction is to explore composing multiple LoRA modules together~\citep{huang2023lorahub}, and we leave it as future work.

\section{Comparison with Concurrent Work}
\label{appendix:concurrent-work} 
CEPE~\cite{yen2024cepe} and SnapKV~\cite{li2024snapkv} are two concurrent work that also proposes methods for context compression. We provide a comparison with them in this section.

\begin{table}[h]
\centering
\resizebox{0.9\linewidth}{!}{\begin{tabular}{llllllcll}
\toprule
Setup & \multicolumn{1}{l}{QuA} & QAS & QM (ROUGE-L) & NQA \\ \midrule
CEPE (official) & 30.2 & 20.5 & 19.6 & 21.9 &  \\
CEPE (reproduced) & 26.40 & 20.54 & 19.38 & 20.33 &  \\
\midrule
\hc LLoCO & \textbf{41.51} & \textbf{29.01} & \textbf{20.92} & \textbf{28.34} \\
\bottomrule
\end{tabular}}
\caption{Comparison of \sys with CEPE~\cite{yen2024cepe}}
\label{table:ceped}
\vspace{-10pt}
\end{table}

\begin{table*}[!ht]
\centering
\resizebox{0.9\linewidth}{!}{
\begin{tabular}{lccc|ccc|ccc|c}
\toprule
& \multicolumn{3}{c|}{SingleDoc} & \multicolumn{3}{c|}{MultiDoc} & \multicolumn{3}{c}{Summarization} \\ \midrule
& NQA & QAS & MFE & HQA & WMQA & MSQ & Gov & QMS & MNews & Avg. \\
\midrule
SnapKV & 20.7 & \textbf{29.3} & \textbf{42.2} & 34.0 & 24.9 & 14.2 & \textbf{28.6} & 23.1 & \textbf{26.5} & 27.1\\
\hc LLoCO & \textbf{23.1} & 26.1 & 26.3 & \textbf{46.2} & \textbf{35.6} & \textbf{27.3} & 17.6 & \textbf{23.4} & 25.0 & \textbf{27.8}\\ \bottomrule
\end{tabular}}
\caption{Comparison between \sys and SnapKV~\cite{li2024snapkv} on LongBench~\citep{longbench-github}}
\label{tab:snapkv}
\end{table*}

\paragraph{Comparison with CEPE} Similar to \sys, CEPE requires training an encoder to generate embeddings for the context. During inference, CEPE uses cross-attention to integrate these embeddings into the original LLM decoder. The main difference is that CEPE applies the encoder at runtime, allowing it to process context in parallel, thus speeding up the inference process. In contrast, \sys performs context encoding offline, retrieving the preprocessed embeddings during inference.

To compare with CEPE, we evaluate their model (based on LLaMA2-7B-Chat) on QuaLITY, Qasper, QMSum, and NarrativeQA. We present two sets of results: the official evaluation performance provided by CEPE, and our reproduced results, obtained by running their publicly available code on these datasets.

As shown in Table~\ref{table:ceped}, LLoCO outperforms CEPE across all datasets. Note that for QMSum dataset, CEPE provided the ROUGE-L score, while \sys provides the geometric mean of ROUGE-1, ROUGE-2, and ROUGE-L scores as our metric. We take the ROUGE-L score of \sys here for a fair comparison.

\paragraph{Comparison with SnapKV}
Compared to \sys and CEPE~\cite{yen2024cepe}, which requires first finetune a encoder beforehand, SnapKV is a KV cache compression approach that is finetuning free. The key insight behind SnapKV is that only a small subset of tokens in the prompt is crucial (those that other tokens attend to), allowing the less important tokens to be discarded.

Since both \sys and SnapKV provide evaluation results on LongBench, we offer a side-by-side comparison of their performance. We extract SnapKV's results using LLaMa2-7B-32k (LongChat) as the base model and compare them with ours. SnapKV provides three settings (1024, 2048, and 4096 tokens), and we use their best results for comparison.

\tbl{tab:snapkv} illustrates that overall, LLoCO and SnapKV show comparable performance across tasks. LLoCO outperforms SnapKV on five datasets, while SnapKV leads on four. Unlike SnapKV, which compresses input tokens to fixed lengths (1024, 2048, 4096), LLoCO applies a consistent 30x compression ratio, enabling the compression of up to 128k tokens into 4096 tokens. This generally results in a higher compression ratio than SnapKV on LongBench.

\end{document}